\documentclass[10pt,twocolumn,letterpaper]{article}

\usepackage{cvpr}
\usepackage{times}
\usepackage{epsfig}
\usepackage{graphicx}
\usepackage{amsmath}
\usepackage{amssymb}
\usepackage{algorithm}
\usepackage{algorithmic}
\usepackage{extarrows}
\usepackage[breaklinks=true,bookmarks=false]{hyperref}

\cvprfinalcopy % *** Uncomment this line for the final submission

\def\cvprPaperID{7913} % *** Enter the CVPR Paper ID here

\ifcvprfinal\fi
\begin{document}
\renewcommand{\thefootnote}{\fnsymbol{footnote}}
%%%%%%%%% TITLE
\title{Hierarchical Clustering with Hard-batch Triplet Loss for Person Re-identification}

\author{Kaiwei Zeng$^{1}$, Munan Ning$^{2}$, Yaohua Wang$^{3}$\footnotemark[1], Yang Guo$^{4}$\footnotemark[1]\\
National University of Defense Technology\\
Changsha,China\\
{\tt\small $^{1}$zengkaiwei1997@gmail.com, $^{2}$munanning@gmail.com, $^{3}$nudtyh@gmail.com, $^{4}$guoyang@nudt.edu.cn}
% \and
% Munan Ning\\
% National University of Defense Technology\\
% China\\
% {\tt\small munanning@gmail.com}
}
\maketitle

\footnotetext[1]{Corresponding author}
\thispagestyle{empty}
%%%%%%%%% ABSTRACT
\begin{abstract}
For clustering-guided fully unsupervised person re-identification (re-ID) methods, the quality of pseudo labels generated by clustering directly decides the model performance. In order to improve the quality of pseudo labels in existing methods, we propose the HCT method which combines \textbf{H}ierarchical \textbf{C}lustering with hard-batch \textbf{T}riplet loss. The key idea of HCT is to make full use of the similarity among samples in the target dataset through hierarchical clustering, reduce the influence of hard examples through hard-batch triplet loss, so as to generate high quality pseudo labels and improve model performance. Specifically, (1) we use hierarchical clustering to generate pseudo labels, (2) we use PK sampling in each iteration to generate a new dataset for training, (3) we conduct training with hard-batch triplet loss and evaluate model performance in each iteration. We evaluate our model on Market-1501 and DukeMTMC-reID. Results show that HCT achieves 56.4\% mAP on Market-1501 and  50.7\% mAP on DukeMTMC-reID which surpasses state-of-the-arts a lot in fully unsupervised re-ID and even better than most unsupervised domain adaptation (UDA) methods which use the labeled source dataset. Code will be released soon on \href{https://github.com/zengkaiwei/HCT}{https://github.com/zengkaiwei/HCT}
\end{abstract}
%%%%%%%%% BODY TEXT
\section{Introduction}
Person re-identification (re-ID) is mainly used to match pictures of the same person that appears in different cameras, which is usually used as an auxiliary method of face recognition to identify pedestrian information. Currently, re-ID has been widely used in the field of security and has been the focus of academic research. With the development of convolutional neural networks (CNN), supervised re-ID \cite{hermans2017defense,li2014deepreid,sun2018beyond,wang2018learning,xu2018attention,chang2018multi,xu2019learning,tian2019person} has achieved excellent performance. However, due to the data deviation in different datasets, the performance of the model trained on the source domain will significantly decline when it is directly transferred to the target domain. Besides, supervised learning requires a large amount of manually annotated data, which is costly in real life. Therefore, supervised re-ID is difficult to meet the requirement of practical application and people tend to focus on unsupervised re-ID. 

\begin{figure}[t]
\begin{center}
\includegraphics[width=0.9\linewidth]{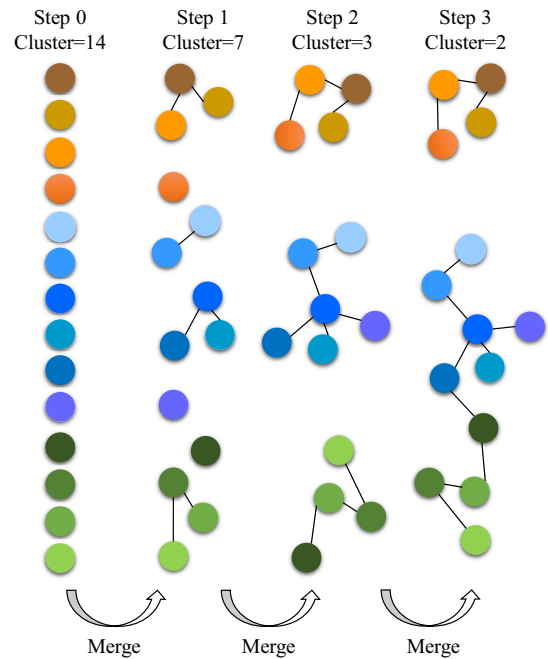}
\end{center}
\caption{Hierarchical clustering. Each circle represents a sample, and the step represents the current merging stage. We use a bottom-up method to merge clusters step by step according to the distance between clusters in the current step.}
\label{fig1}
\end{figure}

Recently, people pay more attention on unsupervised re-ID and achieve good progress. Some works focus on unsupervised domain adaptation (UDA). UDA usually needs manually annotated source data and unlabeled target data. In UDA, some people use GAN to transform the style of images in the source domain to the style of the target domain \cite{deng2018image,wei2018person,zhong2018camera,lv2018unsupervised}. They keep labels unchanged, then they conduct training on generated labeled images. Others focus on the change of images between different cameras and datasets. They identify images by learning differences between the source domain and the target domain \cite{zhong2018generalizing}. Although the expansion of the dataset will generate many reliable data, it is highly dependent on the quality of generated images. Besides, it will also generate some awful images, which will mislead the training and affect model performance. More importantly, these UDA methods only try to reduce differences between the target domain and source domain. However, similarities of images within the target domain are ignored. Besides, UDA methods need a labeled source dataset which still cost a lot.

In recent studies, a fully unsupervised method BUC is proposed \cite{lin2019bottom} and it does not use any manually labeled dataset. BUC only compares the similarity of images in the target dataset and directly use the bottom-up hierarchical clustering to merge samples. BUC merges a fixed number of clusters, updates pseudo labels, and fine-tunes the model step by step until convergence. Finally, it achieves good performance and even surpasses some methods of UDA \cite{deng2018image,fan2018unsupervised,wang2018transferable}. However, the performance of BUC has a significant drop in later merging steps. Because BUC just relies on similarities among samples in merging, it makes BUC difficult to distinguish hard examples, especially in early merging steps when the model is poor. Hard examples
mean those similar samples but have different
identities. Their features are close to each other in high dimensional space so it is difficult to distinguish them by clustering and they will lead to wrong merging. In the later, these wrong merging will generate lots of false pseudo labels which mislead training and result in a decline in performance.

In order to solve these problems and make full use of the similarity of images in the target dataset, we propose HCT, which also a fully unsupervised method just uses the target dataset without any manually annotated labels. The process of hierarchical clustering is shown in Figure \ref{fig1}. In the beginning, we regard each sample as a cluster which has different identities, and then we select a fixed number of clusters for merging in each step according to the distance between clusters. Finally, all clusters will be merged gradually and we set pseudo labels according to clustering results. After clustering, we use hard-batch triplet loss \cite{hermans2017defense} to optimize the model. Hard-batch triplet loss can reduce the distance between similar samples and increase the distance between different samples. It can effectively reduce the influence of hard examples. Specifically, (1) we use hierarchical clustering to merge samples and generate pseudo-labels according to clustering results, (2) we randomly select K instances from P identities (PK sampling) to generate a new dataset for training to meet the need of hard-batch triplet loss, (3) we fine-tune the model and evaluate model performance. We repeat the process of clustering, PK sampling, fine-tuning training, evaluation until the model reaches convergence.

To summarize, our contributions are:
\begin{quote}
\begin{itemize}
\item We propose a fully unsupervised re-ID method HCT. Based on pre-trained ResNet-50\cite{he2016deep} on ImageNet, we directly use pseudo labels generated by hierarchical clustering as supervision to conduct model training on the target dataset without any manually annotated labels. 
\item We use PK sampling to generate a new dataset for training after hierarchical clustering in each iteration. Compared to using the whole datasets, PK sampling meets the need of hard-batch triplet loss\cite{hermans2017defense} which can reduce the influence of hard examples and improve model performance.
\item To correct false pseudo labels, we initialize all pseudo labels at the beginning of each iteration until the quality of pseudo labels stabilizes and the model performance no longer improves.
\item We evaluate our method on Market-1501 and DukeMTMC-reID. Extensive experiments show that our method surpass state-of-the-arts a lot in fully unsupervised re-ID, even better than most UDA methods.
\end{itemize}
\end{quote}

%-------------------------------------------------------------------------
\begin{figure*}
\begin{center}
\includegraphics[width=0.9\linewidth]{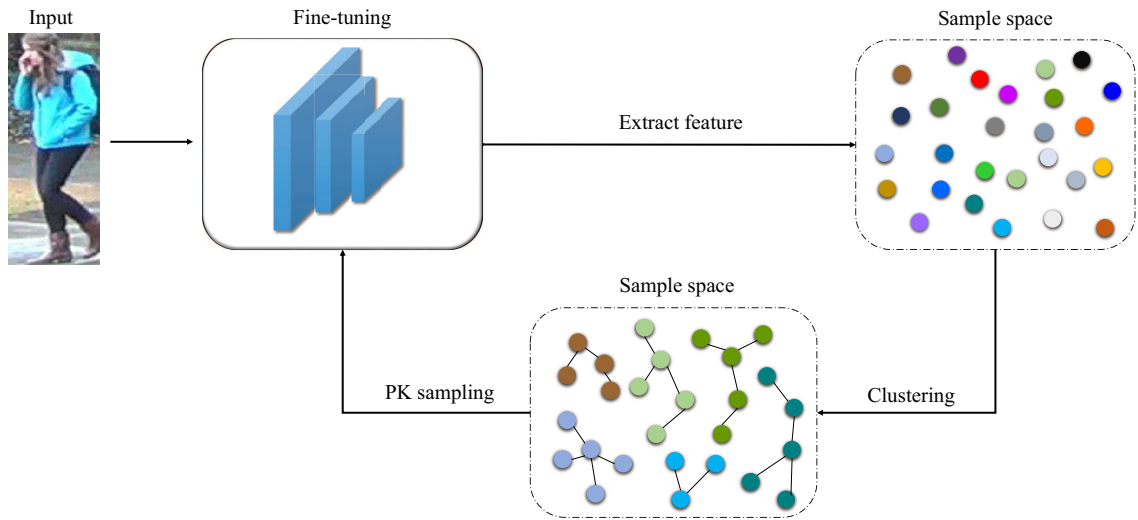}
\end{center}
\caption{The structure of our HCT. Different colors represents different pseudo labels. We use pre-trained ResNet-50 \cite{he2016deep} on ImageNet as our backbone. The input of HCT are unlabeled target images.}
\label{fig2}
\end{figure*}

\section{Related Work}
\subsection{Unsupervised Domain Adaptation Re-ID}
In the past, people tend to use traditional manual features \cite{bazzani2013symmetry,liao2015person} to conduct unsupervised domain adaptation, but the performance on large datasets is usually poor. With the popular of CNN, people begin to apply deep learning to unsupervised domain adaptation. 

Deng et al. put forward SPGAN \cite{deng2018image}. They believe that the main reason for the poor performance of direct transfer is the different camera styles of different datasets. They use CycleGAN \cite{zhong2018camera} to translate images styles from the source domain to the target domain while keeping image labels unchanged. Finally, they perform supervised learning on generated images. Zhong et al. propose ECN \cite{zhong2019invariance}, ECN focuses on exemplar-invariance \cite{wu2018unsupervised,xiao2017joint}, camera-invariance \cite{zhong2018generalizing}, and neighborhood-invariance \cite{choi2018stargan}. Based on these, ECN separately sets triplet loss to increase the distance between different samples and reduce the distance between similar samples. ECN stores samples in the exemplar memory model \cite{santoro2016meta,vinyals2016matching} and sets pseudo labels according to it. Finally, ECN also conducts training according to pseudo labels and get good performance. 

In addition to set pseudo labels as supervision, people also try to use models to learn some auxiliary information to improve generalization ability. Zhong et al. propose HHL \cite{zhong2018generalizing}. HHL improves model performance through camera-invariance and domain connectivity. Xiao et al. propose EANet \cite{huang2018eanet}. EANet proposes Part Aligned Pooling (PAP) and Part Segmentation Constraint (PSC). PAP cuts and aligns images according to the key points of the body posture. PSC enables the model to predict labels of different part about feature map and locate the corresponding position of each part accurately. EANet combines PAP with PSC to make full use of pedestrians pose segmentation information to improve performance. 

Although these methods have achieved some improvements, most of them only focus on the difference between the source domain and the target domain. However, they do not fully explore the similarity of images in the target domain.

\subsection{Clustering-guided Re-ID}
Clustering-guided re-ID is usually trained with pseudo labels generated by clustering, which can be divided into clustering-guided domain adaptation and clustering-guided fully unsupervised re-ID. 

For clustering-guided domain adaptation, Hehe et al. \cite{fan2018unsupervised} propose PUL. PUL gets the pre-trained model through training on the labeled source dataset, then uses CNN to fine-tune the model and uses K-means to cluster samples. At the beginning of training, PUL only selects a part of reliable samples which close to the clustering centroid for training to avoid falling into local optimum. As the model becomes better, more samples will be selected. This strategy effectively promotes the convergence of the model and improves performance. However, K-means is very sensitive to the k value. Besides, as a a partition based clustering method, clustering centroids are easily dragged by outliers, it will generate lots of false pseudo labels which seriously affect the optimization of the model and ultimately limit model performance. 

For clustering-guided fully unsupervised re-ID, Lin et al. propose BUC \cite{lin2019bottom}. BUC does not use any labeled source data, only use unlabeled target data and pre-trained model on ImageNet instead of other re-ID dataset. BUC extracts image features with CNN, then merges a fixed number of clusters according to the distance between clusters in each step. After merging, BUC fine-tunes the model with generated pseudo labels, repeats the progress of merging and fine-tuning until the model performance no longer improves. However, the performance of BUC has a significant drop in later merging steps. That is due to the poor pre-trained model in the beginning and some hard examples in the target dataset. BUC cannot solve the problem of false pseudo labels, which will affect the optimization of the model. These false pseudo labels have a superposition effect in later merging steps and result in a significant performance drop in the end. In this paper, we aim to further improve the quality of pseudo labels and get better performance than these methods.

\section{Our Method}
\subsection{Hierarchical Clustering with Hard-batch Triplet Loss}

Our network structure is shown in Figure \ref{fig2}. The model is mainly divided into three stages: hierarchical clustering, PK sampling, and fine-tuning training. We extract image features to form a sample space and cluster samples step by step according to the bottom-up hierarchical clustering in Figure \ref{fig1}. After hierarchical clustering, we label samples in the same cluster with the same pseudo label. Finally, we use PK sampling to generate a new dataset for training according to clustering results. Our goal is to explore similarities among images in the target dataset through hierarchical clustering, distinguish hard examples through hard-batch triplet loss, and generate pseudo labels to guide model training in the end. Compared to other methods, our HCT can further improve the quality of pseudo labels and finally get better model performance. 

For a dataset $X =\left\{x_{1},x_{2},\cdots,x_{N}\right\}$, we will have manually annotated labels $Y =\left\{y_{1},y_{2},\cdots,y_{n}\right\}$ in supervised learning, so we can directly use cross-entropy loss to optimize the model. However, we do not have any manually annotated labels in fully unsupervised re-ID, so we need to generate pseudo labels as supervision instead of using manually annotated labels. Although hierarchical clustering can fully explore the similarity of samples, build the underlying structure through a bottom-up clustering and generate some good pseudo labels. But due to the deficiency of hierarchical clustering, this strategy cannot effectively distinguish hard examples and will generate lots of false pseudo labels in merging. These false pseudo labels will mislead the optimization of model and limit model performance. 

In order to solve this problem, HCT uses hard-batch triplet loss with PK sampling to reduce the distance between similar samples and increase the distance between different samples, which can better distinguish hard examples. Besides, we will initialize all pseudo labels at the beginning of each iteration so that we are able to correct all false pseudo labels generated in the previous iteration. Theoretically, as pseudo labels of hierarchical clustering are approaching manually annotated labels step by step, the model performance is approaching the baseline. Baseline represents the supervised learning method of hard-batch triplet loss.

\subsection{Distance Measurement}
For all clustering-guided re-ID \cite{lin2019bottom,fan2018unsupervised,song2018unsupervised}, the quality of pseudo labels generated by clusters  directly determines the performance of the model. For hierarchical clustering, the distance measurement method used in the merging stage decide how we choose clusters to merge and finally affects the clustering result and pseudo labels. 

BUC \cite{lin2019bottom} uses the minimum distance as the distance measurement in the merging stage. Minimum distance only calculates one pair of the nearest pairwise distance in two clusters. That is not a good method because it ignores other samples in clusters. Especially when there are lots of samples in a cluster, minimum distance is easily influenced by outliers and finally results in wrong merging and false pseudo labels. To improve the distance measurement and finally get a better result, we should consider the pairwise distance of all the samples in two clusters. 

In HCT, we use euclidean distance to measure the distance between each sample. Then, according to the unweighted average linkage clustering (unweighted pair-group method with arithmetic means, UPGMA) \cite{sokal1958statistical}, we define the distance between clusters as:
\begin{equation}
\label{equ:E1}
D_{ab}=\frac{1}{n_{a} n_{b}} \sum_{i \in {C}_{a}, j \in {C}_{b}}{D}\left({C}_{a_{i}}, {C}_{b_{j}}\right)       \end{equation}
where ${C}_{a_{i}}$,${C}_{b_{j}}$ are two samples in the cluster ${C}_{a}$,${C}_{b}$ respectively. $n_{a}$,$n_{b}$ represent the number of samples in ${C}_{a}$,${C}_{b}$, $D(\cdot)$ means the euclidean distance. UPGMA takes into account all the pairwise distance between two clusters and each pairwise distance has the same weight. It effectively reduces the influence of outliers in sample space, promote more rational merging and finally get better results compared to other distance measurement according to discussion in \cite{ding2019towards}.

\subsection{Loss Function}
Hard-batch triplet loss \cite{hermans2017defense} is proposed to mine the relationship between $anchor$ with $positive$ $sample$ and $negative$ $sample$, which can reduce the distance between similar samples and increase the distance between different samples. In order to use hard-batch triplet loss in HCT, we use PK sampling to generate a new dataset for training. Specifically, We randomly select K instances from P identities for each mini-batch (batchsize = P$\times$K). So our loss is defined as:

\begin{align}
L_{t r i p l e t}=\sum_{i=1}^{P} \sum_{a=1}^{K}\Big[m +\overbrace{\max _{p=1 \ldots K} D\left(x_{a}^{i},x_{p}^{j}\right)}^{\text {hardest positive}} \nonumber \\ 
-\underbrace{\min _{j=1 \ldots P \atop n=1 \ldots N} D\left(x_{a}^{i},x_{n}^{j}\right)}_{\text {hardest negative }} \Big] 
\end{align}
where $x_{a}^{i}$ is the $anchor$, $x_{p}^{j}$ is the $positive$ $sample$ which has the same identity as $x_{a}^{i}$, $x_{n}^{j}$ is the $negative$ $sample$ which identity is different from $x_{a}^{i}$. $D(\cdot)$ means the euclidean distance and $m$ is the hyperparameter $margin$ in hard-batch triplet loss. Hard-batch triplet loss makes sure that give an anchor $x_{a}^{i}$, $x_{p}^{j}$ is closer to $x_{a}^{i}$ than $x_{n}^{j}$. As a result, samples which have the same identity will be closer to each other than other samples which have different identities. In other words, these samples will form a cluster gradually in high dimensional space. So we can use hard-batch triplet loss to distinguish hard examples, promote better clustering, and improve model performance.

\subsection{Model Update}
As shown in the algorithm, we use pre-trained ResNet-50 \cite{he2016deep} on ImageNet for training. For each iteration, at the beginning of hierarchical clustering, we regard $N$ samples as $N$ different identities and initialize all pseudo labels. We set a hyperparameter $mp$ to control the speed of the merging and a hyperparameter $s$ represents total merging steps of hierarchical clustering, $m =n\times mp$ represents the number of clusters merged in each step. We calculate all pairwise distance between samples in the target dataset and generate a $n\times n$ distance matrix $dist$. According to $dist$ and UMPGA distance measurement in Eq.(\ref{equ:E1}), we generate a $c\times c$ distance matrix $\boldsymbol{D}$, $\boldsymbol{D}$ represents the distance between clusters in each step, $c$ represents the current number of clusters. We will merge $m$ pairs of nearest clusters in each step until the $s$-$th$ step and generate pseudo labels according to the clustering result. Specifically, we regard samples in the same cluster have the same pseudo labels. Then we use PK sampling to generate a new dataset as the input of CNN, we conduct fine-tuning training with the new dataset and evaluate model performance in the end. We regard hierarchical clustering, PK sampling, fine-tuning training and evaluation as one iteration. We iterate the model until the performance no longer improves.

%------------------------------------------------------------------------

\begin{algorithm}[ht]
\caption{HCT Algorithm}
\begin{algorithmic}[1]
\REQUIRE ~~\\
\quad Input $X=\left\{x_{1},x_{2},\cdots,x_{N}\right\}$; \\
\quad Merging percent $mp \in(0,1)$; \\
\quad Merging step $s$; \\
\quad Iteration $t$. \\
\ENSURE ~~\\
\quad Best model $f\left(\boldsymbol{w}, x_{i}\right)$.
\STATE Initialize: \\
\qquad sample number $n=N$,\\
\qquad cluster number $c=n$,\\
\qquad merging number $m=n\times mp$,\\
\qquad iteration $iter = 0$,\\
\qquad merging step $step = 0$.
\WHILE {$iter<t$}
\STATE Initialize pseudo labels: $Y=\left\{y_{i}=i\right\}_{i=1}^{N}$;
\STATE Extract features, calculate the pairwise distance between each sample, and generate a $n\times n$ distance matrix $dist$;
\WHILE {$step<s$}
\STATE Calculate distance between each cluster according to Eq.(\ref{equ:E1}), generate a $c\times c$ distance matrix $\boldsymbol{D}$;
\STATE Select clusters to merge according to $\boldsymbol{D}$ and start to merge clusters: \\
\qquad $c=c-m$;
\STATE Update $Y$ with new pseudo labels:\\
\qquad $Y=\left\{y_{i}=j, \quad \forall x_{i} \in {C}_{j}\right\}_{i=1}^{N}$; \\
\qquad $step = step+1$;
\ENDWHILE
\STATE Generate a new dataset with PK sampling according to $Y$; \\
\STATE Fine-tuning model with the new dataset according to hard-batch triplet loss; \\
\STATE Evaluate model performance; \\
\IF{$mAP_{i}>mAP_{best}$}
\STATE $mAP_{best}=mAP_{i}$;
\STATE Best model $f\left(\boldsymbol{w}, x_{i}\right)$;
\ENDIF 
\STATE $iter = iter+1$;
\ENDWHILE 
\end{algorithmic}
\end{algorithm}

\section{Experiment}
\subsection{Datasets}
\textbf{Market-1501}
Market1501 \cite{zheng2015scalable} includes 32,668 images of 1501 pedestrians captured by 6 cameras. Each pedestrian is captured by at least two cameras. Market1501 can be divided into a training set which contains 12,936 images of 751 people and a test set which contains 19,732 images of 750 people.

\textbf{DukeMTMC-reID}
DukeMTMC-reID \cite{zheng2017unlabeled} is a subset of pedestrian re-identification dataset DukeMTMC \cite{ristani2016performance}. DukeMTMC contains a 85 minutes high-resolution video, which is collected from eight different cameras. DukeMTMC-reID consists of 36411 labelled images belonging to 1404 identities which contains 16,522 images for training, 2,228 images for query, and 17,661 images for gallery.

% \textbf{MSMT17}
% MSMT17 \cite{wei2018person} contains 126,441 bounding boxes of 4,101 identities taken by 12 outdoor cameras and 3 indoor cameras in four days. The training set includes 1041 pedestrians with 32,621 bounding boxes, the test set include 3060 pedestrians with 93,820 bounding boxes. For the test set, 11,659 bounding boxes are randomly selected as query, the other 82,161 bounding boxes are gallery. MSMT17 dataset has a large scale, complex scene, and covers multiple time periods. At the same time, RCNN \cite{ren2015faster} is used to detect pedestrian bounding boxes. MSMT17 is one of the most challenging re-ID datasets at present.
%-------------------------------------------------------------------------
\begin{table*}[t]
\center
\begin{tabular}{lllcccccccc}
\hline
\multicolumn{1}{|l|}{\begin{tabular}[c]{@{}l@{}} \\ Methods\end{tabular}}  & \multicolumn{1}{l|}{\begin{tabular}[c]{@{}l@{}} \\ Labels\end{tabular}} & \multicolumn{4}{c|}{Market-1501} & \multicolumn{4}{c|}{DukeMTMC-reID} \\ \cline{3-10} 
\multicolumn{1}{|l|}{} & \multicolumn{1}{l|}{} & \multicolumn{1}{c|}{rank-1} & \multicolumn{1}{c|}{rank-5} & \multicolumn{1}{c|}{rank-10} & \multicolumn{1}{c|}{mAP} & \multicolumn{1}{c|}{rank-1} & \multicolumn{1}{c|}{rank-5} & \multicolumn{1}{c|}{rank-10} & \multicolumn{1}{c|}{mAP} \\ \hline
\multicolumn{1}{|l|}{Baseline\cite{song2018unsupervised}} & \multicolumn{1}{l|}{Supervised} & \multicolumn{1}{c|}{91.6} & \multicolumn{1}{c|}{-} & \multicolumn{1}{c|}{-} & \multicolumn{1}{c|}{78.2} & \multicolumn{1}{c|}{80.8} & \multicolumn{1}{c|}{-} & \multicolumn{1}{c|}{-} & \multicolumn{1}{c|}{65.4} \\ 
\multicolumn{1}{|l|}{Direct transfer} &\multicolumn{1}{l|}{None} & \multicolumn{1}{c|}{11.1} & \multicolumn{1}{c|}{22.1} & \multicolumn{1}{c|}{28.6} & \multicolumn{1}{c|}{3.5} & \multicolumn{1}{c|}{8.6} & \multicolumn{1}{c|}{16.4} & \multicolumn{1}{c|}{21.0} & \multicolumn{1}{c|}{3.0} \\ 
\multicolumn{1}{|l|}{HCT} & \multicolumn{1}{l|}{None} & \multicolumn{1}{c|}{80.0} & \multicolumn{1}{c|}{91.6} & \multicolumn{1}{c|}{95.2} & \multicolumn{1}{c|}{56.4} & \multicolumn{1}{c|}{69.6} & \multicolumn{1}{c|}{83.4} & \multicolumn{1}{c|}{87.4} & \multicolumn{1}{c|}{50.7}\\ \hline
\end{tabular}
\caption{Comparison with baseline and direct transfer on Market-1501 and DukeMTMC-reID . "Baseline" means supervised learning method about hard-batch triplet loss. "Direct transfer" means directly use pre-trained ResNet-50 on ImageNet to evaluate without any fine-tuning. The label column lists the type of supervision used by the method. "Supervised" means supervised learning, "None" denotes no any manually annotated labels are used, which is fully unsupervised learning.}
\label{table1}
\end{table*}

\begin{table*}[t]
\center
\begin{tabular}{lllcccccccc}
\hline
\multicolumn{1}{|l|}{\begin{tabular}[c]{@{}l@{}} \\ Methods\end{tabular}}  & \multicolumn{1}{l|}{\begin{tabular}[c]{@{}l@{}} \\ Labels\end{tabular}} & \multicolumn{4}{c|}{Market-1501} & \multicolumn{4}{c|}{ DukeMTMC-reID} \\ \cline{3-10} 
\multicolumn{1}{|l|}{} & \multicolumn{1}{l|}{} & \multicolumn{1}{c|}{rank-1} & \multicolumn{1}{c|}{rank-5} & \multicolumn{1}{c|}{rank-10} & \multicolumn{1}{c|}{mAP} & \multicolumn{1}{c|}{rank-1} & \multicolumn{1}{c|}{rank-5} & \multicolumn{1}{c|}{rank-10} & \multicolumn{1}{c|}{mAP} \\ \hline
\multicolumn{1}{|l|}{UMDL\cite{peng2016unsupervised}} & \multicolumn{1}{l|}{Transfer} & \multicolumn{1}{c|}{34.5} & \multicolumn{1}{c|}{52.6} & \multicolumn{1}{c|}{59.6} & \multicolumn{1}{c|}{12.4} & \multicolumn{1}{c|}{18.5} & \multicolumn{1}{c|}{31.4} & \multicolumn{1}{c|}{37.4} & \multicolumn{1}{c|}{7.3} \\ 
\multicolumn{1}{|l|}{OIM\cite{xiao2017joint}*} & \multicolumn{1}{l|}{None} & \multicolumn{1}{c|}{38.0} & \multicolumn{1}{c|}{58.0} & \multicolumn{1}{c|}{66.4} & \multicolumn{1}{c|}{14.0} & \multicolumn{1}{c|}{24.5} & \multicolumn{1}{c|}{38.8} & \multicolumn{1}{c|}{46.0} & \multicolumn{1}{c|}{11.3} \\ 
\multicolumn{1}{|l|}{PUL\cite{fan2018unsupervised}} &\multicolumn{1}{l|}{Transfer} & \multicolumn{1}{c|}{45.5} & \multicolumn{1}{c|}{60.7} & \multicolumn{1}{c|}{66.7} & \multicolumn{1}{c|}{20.5} & \multicolumn{1}{c|}{30.0} & \multicolumn{1}{c|}{43.4} & \multicolumn{1}{c|}{48.5} & \multicolumn{1}{c|}{16.4} \\ 
\multicolumn{1}{|l|}{SPGAN\cite{deng2018image}} & \multicolumn{1}{l|}{Transfer} & \multicolumn{1}{c|}{51.5} & \multicolumn{1}{c|}{70.0} & \multicolumn{1}{c|}{76.8} & \multicolumn{1}{c|}{22.8} & \multicolumn{1}{c|}{41.1} & \multicolumn{1}{c|}{56.6} & \multicolumn{1}{c|}{63.0} & \multicolumn{1}{c|}{22.3} \\ 
\multicolumn{1}{|l|}{TJ-AIDL\cite{wang2018transferable}} &  \multicolumn{1}{l|}{Transfer} & \multicolumn{1}{c|}{58.2} & \multicolumn{1}{c|}{74.8} & \multicolumn{1}{c|}{81.1} & \multicolumn{1}{c|}{26.5} & \multicolumn{1}{c|}{44.3} & \multicolumn{1}{c|}{59.6} & \multicolumn{1}{c|}{65.0} & \multicolumn{1}{c|}{23.0} \\ 
\multicolumn{1}{|l|}{HHL\cite{zhong2018generalizing}} & \multicolumn{1}{l|}{Transfer} & \multicolumn{1}{c|}{62.2} & \multicolumn{1}{c|}{78.8} & \multicolumn{1}{c|}{84.0} & \multicolumn{1}{c|}{31.4} & \multicolumn{1}{c|}{46.9} & \multicolumn{1}{c|}{61.0} & \multicolumn{1}{c|}{66.7} & \multicolumn{1}{c|}{27.2} \\ 
\multicolumn{1}{|l|}{BUC\cite{lin2019bottom}} & \multicolumn{1}{l|}{None} & \multicolumn{1}{c|}{66.2} & \multicolumn{1}{c|}{79.6} & \multicolumn{1}{c|}{84.5} & \multicolumn{1}{c|}{38.3} & \multicolumn{1}{c|}{47.4} & \multicolumn{1}{c|}{62.6} & \multicolumn{1}{c|}{68.4} & \multicolumn{1}{c|}{27.5} \\ 
\multicolumn{1}{|l|}{ARN\cite{li2018adaptation}} &\multicolumn{1}{l|}{Transfer} & \multicolumn{1}{c|}{70.3} & \multicolumn{1}{c|}{80.4} & \multicolumn{1}{c|}{86.3} & \multicolumn{1}{c|}{39.4} & \multicolumn{1}{c|}{60.2} & \multicolumn{1}{c|}{73.9} & \multicolumn{1}{c|}{79.5} & \multicolumn{1}{c|}{33.4} \\ 
\multicolumn{1}{|l|}{MAR\cite{yu2019unsupervised}} & \multicolumn{1}{l|}{Transfer} & \multicolumn{1}{c|}{67.7} & \multicolumn{1}{c|}{81.9} & \multicolumn{1}{c|}{-} & \multicolumn{1}{c|}{40.0} & \multicolumn{1}{c|}{67.1} & \multicolumn{1}{c|}{79.8} & \multicolumn{1}{c|}{-} & \multicolumn{1}{c|}{48.0} \\ 
\multicolumn{1}{|l|}{ECN\cite{zhong2019invariance}}  & \multicolumn{1}{l|}{Transfer} & \multicolumn{1}{c|}{75.1} & \multicolumn{1}{c|}{87.6} & \multicolumn{1}{c|}{91.6} & \multicolumn{1}{c|}{43.0} & \multicolumn{1}{c|}{63.3} & \multicolumn{1}{c|}{75.8} & \multicolumn{1}{c|}{80.4} & \multicolumn{1}{c|}{40.4} \\ 
\multicolumn{1}{|l|}{EANet\cite{huang2018eanet}}  & \multicolumn{1}{l|}{Transfer} & \multicolumn{1}{c|}{78.0} & \multicolumn{1}{c|}{-} & \multicolumn{1}{c|}{-} & \multicolumn{1}{c|}{51.6} & \multicolumn{1}{c|}{67.7} & \multicolumn{1}{c|}{-} & \multicolumn{1}{c|}{-} & \multicolumn{1}{c|}{48.0} \\ 
\multicolumn{1}{|l|}{Theory\cite{song2018unsupervised}} & \multicolumn{1}{l|}{Transfer} & \multicolumn{1}{c|}{75.8} & \multicolumn{1}{c|}{85.9} & \multicolumn{1}{c|}{93.2} & \multicolumn{1}{c|}{53.7} & \multicolumn{1}{c|}{68.4} & \multicolumn{1}{c|}{80.1} & \multicolumn{1}{c|}{83.5} & \multicolumn{1}{c|}{49.0} \\ 
\multicolumn{1}{|l|}{HCT} & \multicolumn{1}{l|}{None} & \multicolumn{1}{c|}{\textbf{80.0}} & \multicolumn{1}{c|}{\textbf{91.6}} & \multicolumn{1}{c|}{\textbf{95.2}} & \multicolumn{1}{c|}{\textbf{56.4}} & \multicolumn{1}{c|}{\textbf{69.6}} & \multicolumn{1}{c|}{\textbf{83.4}} & \multicolumn{1}{c|}{\textbf{87.4}} & \multicolumn{1}{c|}{\textbf{50.7}} \\ \hline
\end{tabular}
\caption{Comparison with other unsupervised methods. The label column lists the type of supervision used by the method. "Transfer" means uses an manually annotated source dataset, which is UDA method. "*" means results are reported by \cite{lin2019bottom}. Results that surpass all competing methods are \textbf{bold}.}
\label{table2}
\end{table*}

\subsection{Implementation Details}
\textbf{HCT Training Setting}
We directly use pre-trained ResNet \cite{he2016deep} on ImageNet as our backbone. After clustering, we randomly selected $P = 16$ identities and $K = 4$ images to generate a new train dataset, so $batch size = P\times K = 64$. During the training, we adjust the size of the input image to $256 \times 128$, we also use random cropping, flipping
and random erasing for data augmentation \cite{zhong2017random}. We use SGD to optimize the model and set a momentum \cite{sutskever2013importance} of 0.9 without dampening. The learning rate is $6 \times10^{-5}$, the weight decay is $0.0005$, iteration is 20, and $margin$ is $0.5$ in hard-batch triplet loss. In Market-1501, merging percent $mp$ is 0.07, merging step $s$ is 13, epoch is 60. 
Note that the model is easily to overfit and will have a significant drop in the later iteration, we adopt an early stop strategy to get best performance.

\textbf{Evaluating Setting}
We use the single-shot setting \cite{sun2018beyond} in all experiments. In evaluation, for an image in query, we calculate cosine distance with all gallery images and then sort it as the result. We use the mean average precision (mAP) \cite{zheng2015scalable} and the rank-$k$ accuracy to evaluate the performance of the model. Rank-$k$ emphasizes the accuracy, it means the query picture has the match in the top-$k$ list. Beside, mAP is computed from the Cumulated Matching Characteristics (CMC) \cite{gray2007evaluating}. CMC curve shows the probability that a query has the match in different size of lists. Given a single query, the Average Precision (AP) is computed according to its precision-recall curve, the mAP is the mean of AP.

\subsection{Ablation Study}

\textbf{Comparison with Baseline and Direct Transfer}
In order to reflect the effect of our HCT, we compare HCT with a supervised learning method about hard-batch triplet loss and direct transfer from pre-trained ResNet-50 on ImageNet. Our results are reported in Table \ref{table1}.
The results for direct transfer and baseline represent the  floor and upper limit of model performance. Theoretically, when the quality of our pseudo labels approach to manually annotated labels, HCT will gradually approach the baseline.

We can see that the performance of direct transfer is very poor, only get $3.5\%$ mAP on Market-1501 and $3.0\%$ mAP on DukeMTMC-reID. That is because the model is pre-trained on ImageNet for a classification task which is completely different from re-ID task. HCT outperforms the direct transfer method by $52.9\%$ mAP on Market-1501 and $47.7\%$ mAP on DukeMTMC-reID. That is only less than supervised method baseline $21.8\%$ mAP and $14.7\%$ mAP respectively, which indicates that the quality of pseudo labels generated by HCT is very high, so our model performance is good.

\begin{table}[t]
\center
\begin{tabular}{lllcccccccc}
\hline
\multicolumn{1}{|l|}{\begin{tabular}[c]{@{}l@{}} \\ Merge step \end{tabular}}  &  \multicolumn{4}{c|}{ Market-1501} \\ \cline{2-5} 
\multicolumn{1}{|l|}{} & \multicolumn{1}{c|}{IDs}  & \multicolumn{1}{c|}{epoch} & \multicolumn{1}{c|}{rank-1} & \multicolumn{1}{c|}{mAP} \\ \hline
\multicolumn{1}{|c|}{s = 12} & \multicolumn{1}{c|}{2069} & \multicolumn{1}{c|}{15} & \multicolumn{1}{c|}{72.2} & \multicolumn{1}{c|}{46.2} \\ 
\multicolumn{1}{|c|}{s = 13} & \multicolumn{1}{c|}{1171}& \multicolumn{1}{c|}{60} & \multicolumn{1}{c|}{\textbf{80.0}} & \multicolumn{1}{c|}{\textbf{56.4}} \\ 
\multicolumn{1}{|c|}{s = 14} & \multicolumn{1}{c|}{258}& \multicolumn{1}{c|}{300} & \multicolumn{1}{c|}{$\times$} & \multicolumn{1}{c|}{$\times$} \\
\hline
\end{tabular}
\caption{Performance comparison with different merging steps on Market-1501. "IDs" means identities number, it also represent the number of clusters after hierarchical clustering. "Epoch" means the training epoch in each iteration. "$\times$" means the model is difficult to converge.}
\label{table3}
\end{table}

\textbf{Effectiveness of HCT}
As shown in Table \ref{table2}, we compare our HCT with other unsupervised methods. On Market-1501, we obtain \textbf{rank-1 =80.0\%, mAP =56.4\%}. On DukeMTMC-reID, we obtain \textbf{rank-1 =69.6\%, mAP =50.7\%}. HCT not only surpasses other fully unsupervised methods a lot, but also better than many UDA methods. Note that we do not use any manually labeled data for training, we just use unlabeled target data. Results indicates the importance of fully exploring the similarity of the samples in the target domain. Besides, it also proves that hard-batch triplet loss can effectively reduce the influence of hard examples, further improve the quality of pseudo labels, and get better performance.

\textbf{Comparison with Different Merging Steps}
In hierarchical clustering, merging step $s$ controls the termination of merging, determines clusters number, and finally affects the quality of pseudo labels. In order to get the best performance, we set $mp=0.07$ and evaluate the impact of different $s$ on Market-1501. Our results are reported in Table \ref{table3}. When we set $s=14$, we find the model is difficult to converge even if we set the training epoch very high. Market-1501 have 751 IDs in the training set, but now HCT only have 258 IDs. We believe that in the final merge step, hierarchical clustering will generate lots of awful clusters and false pseudo labels that cannot be optimized. So we should adopt an early stop strategy in hierarchical clustering. However, too small $s$ means too many IDs number of pseudo labels which will also cause problems. When we set $s=12$, we can see a significant decline on performance. So too early stop will reduces the performance of the model. Besides, we have to decrease the training epoch to 15, because we find a large epoch easily leads to overfitting when $s$ is too small. Finally, we get the best performance when we set $s=13$. 

\begin{table}[t]
\center
\begin{tabular}{lllcccccccc}
\hline
\multicolumn{1}{|l|}{\begin{tabular}[c]{@{}l@{}} \\ Merge percent \end{tabular}}  &  \multicolumn{4}{c|}{ Market-1501} \\ \cline{2-5} 
\multicolumn{1}{|l|}{} & \multicolumn{1}{c|}{rank-1} &\multicolumn{1}{c|}{rank-5} &\multicolumn{1}{c|}{rank-10} & \multicolumn{1}{c|}{mAP} \\ \hline
\multicolumn{1}{|c|}{mp = 0.04} & \multicolumn{1}{c|}{79.6} & \multicolumn{1}{c|}{90.9} &\multicolumn{1}{c|}{94.6} & \multicolumn{1}{c|}{55.3}  \\
\multicolumn{1}{|c|}{mp = 0.05} &\multicolumn{1}{c|}{78.7} &\multicolumn{1}{c|}{91.1} & \multicolumn{1}{c|}{94.6}& \multicolumn{1}{c|}{55.0}  \\ 
\multicolumn{1}{|c|}{mp = 0.06} &\multicolumn{1}{c|}{78.1} &\multicolumn{1}{c|}{91.1} & \multicolumn{1}{c|}{94.2}& \multicolumn{1}{c|}{54.3}  \\ 
\multicolumn{1}{|c|}{mp = 0.07} &\multicolumn{1}{c|}{\textbf{80.0}} &\multicolumn{1}{c|}{\textbf{91.6}} & \multicolumn{1}{c|}{\textbf{95.2}}& \multicolumn{1}{c|}{\textbf{56.4}}\\ 
\multicolumn{1}{|c|}{mp = 0.08} &\multicolumn{1}{c|}{77.0} &\multicolumn{1}{c|}{90.4} & \multicolumn{1}{c|}{94.1}& \multicolumn{1}{c|}{53.0} \\ 
\multicolumn{1}{|c|}{mp = 0.09} &\multicolumn{1}{c|}{77.9} &\multicolumn{1}{c|}{90.8} & \multicolumn{1}{c|}{94.2}& \multicolumn{1}{c|}{54.6}  \\ 
\multicolumn{1}{|c|}{mp = 0.1} & \multicolumn{1}{c|}{77.4}& \multicolumn{1}{c|}{90.9}& \multicolumn{1}{c|}{94.6}& \multicolumn{1}{c|}{53.7} \\ 
\hline
\end{tabular}
\caption{Performance comparison with different merging percents on Market-1501.}
\label{table4}
\end{table}

\textbf{Comparison with Different Merging Percents}
In hierarchical clustering, merging percent $mp$ controls the speed of merging. It decides the number of clusters merged in each step and finally affects generated pseudo labels. In order to get the best performance and evaluate the influence of $mp$, we evaluate different $mp$ values on Market-1501. Based on discussion above, we adopt an early stop strategy in our experiments for the setting of $s$ in all experiments. In Table \ref{table4}, we can see that when we set $mp=0.07$, we get the best performance. We believe that both too many and too few merging in each step will result in a decline of cluster quality.  Besides, compared to change merging step $s$, changing merging percent $mp$ only causes a slight change in performance. 

\begin{figure*}
\begin{center}
\includegraphics[width=0.9\linewidth]{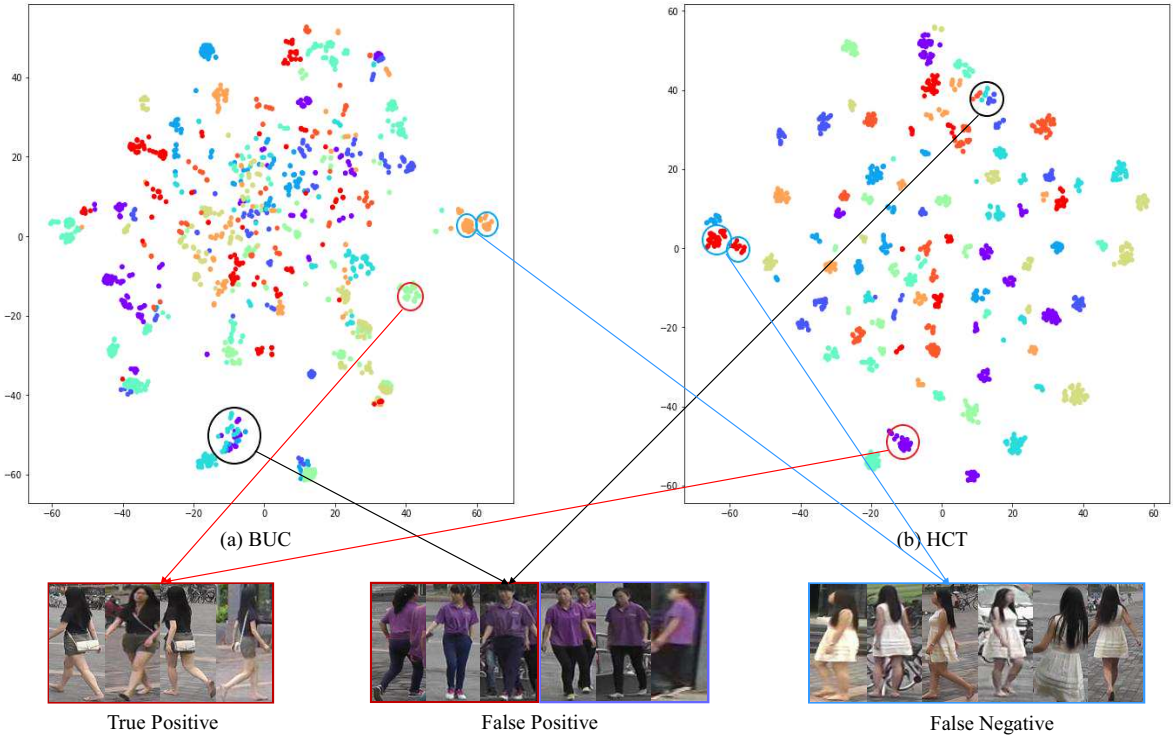}
\end{center}
\caption{T-SNE visualization of the feature representation on a subset of Market-1501 about BUC (100 identities and 1747images) and HCT (100 identities and 1656 images). Samples with the same color represent them have same real labels. $True$ $Positive$ means correct pseudo labels generated by model. $False$ $Positive$ means model generate the same pseudo label for images that belong to different identities in fact. $False$ $Negative$ means model generate different pseudo labels for images that belong to the same identity in fact. Both $False$ $Positive$ and $False$ $Negative$ will generate false pseudo labels which reduce the model performance.}
\label{fig3}
\end{figure*}

\textbf{Qualitative Analysis of T-SNE Visualization}
As shown in Figure \ref{fig3}, we can see that BUC cannot effectively distinguish hard examples, so there are lots of $False$ $Positive$ samples in clusters. These $False$ $Positive$ samples are close to each other in high dimensional space and easily result in wrong merging in hierarchical clustering. Besides, the distribution of clustering results is dispersing and will generate many $False$ $Negative$ samples. Different from hard examples, these $False$ $Negative$ samples belong to the same identity. But they are not very close to each other in high dimensional space, so we cannot effectively use hierarchical clustering to merge them into one cluster. Our method HCT solves these problems and get better performance. We can see that HCT can promote more compact clustering, so the number of $False$ $Negative$ samples is greatly reduced. Besides, HCT can effectively distinguish hard examples, so the number of $False$ $Positive$ samples is also greatly reduced. These results illustrate the effectiveness of hard-batch triplet loss and the high quality of pseudo labels generated by HCT. In general, due to the significant improvement of clustering result, HCT surpasses a lot than other unsupervised methods.

\section{Conclusion}
In this paper, we propose a fully unsupervised re-ID method HCT. HCT directly use unlabeled dataset without using any manually annotated labels for training. We make full use of similarities between images in the target dataset through hierarchical clustering. We also effectively reduced the influence of hard examples in training by PK sampling and hard-batch triplet loss. Besides, we further improve the quality of generated pseudo labels through initializing pseudo labels and training alternately. Finally, as the quality of pseudo labels gradually improving, our model performance are improving step by step. Extensive experiments prove that HCT surpasses state-of-the-arts in fully unsupervised methods by a large margin, even better than most UDA methods.

\section{Acknowledgements}
We thank the reviewers for their feedback. We thank our group members for feedback and the stimulating intellectual environment they provide. This research was supported by The Science and Technology Planning Project of Hunan Province (No.2019RS2027) and National Key Research and Development Program of China (No.2018YFB0204301).

{\small
\bibliographystyle{ieee_fullname}
\bibliography{egbib}
}

\end{document}